 \documentclass[journal,comsoc]{IEEEtran}

\usepackage[T1]{fontenc}

\usepackage{cite}

\ifCLASSINFOpdf
\else
\fi

\usepackage{algorithmic}
\usepackage{array}

\ifCLASSOPTIONcompsoc
  \usepackage[caption=false,font=normalsize,labelfont=sf,textfont=sf]{subfig}
\else
  \usepackage[caption=false,font=footnotesize]{subfig}
\fi
\usepackage{url}

\usepackage{textcomp}
\usepackage{graphicx}
\usepackage{textcomp}
\usepackage{comment}

\usepackage{amssymb}
\usepackage{aurical}
\usepackage[T1]{fontenc}
\usepackage[utf8]{inputenc}
\usepackage[normalem]{ulem}
\usepackage{array}
\usepackage{fancyhdr}
\usepackage{xcolor}
\usepackage{amsthm}

\newcommand{\myuuline}[1]{{#1}}
\newcommand{\myuline}[1]{{#1}}
\newcommand{\muline}[1]{{#1}}

\begin{document}

 \title{
 The Internet of Senses:\\ Building on Semantic Communications  and Edge Intelligence}

\author{Roghayeh~Joda,~\IEEEmembership{Senior Member,~IEEE},         Medhat~Elsayed, Hatem~Abou-zeid, Ramy~Atawia, Akram~Bin Sediq, Gary~Boudreau,~\IEEEmembership{Senior Member,~IEEE}, Melike~Erol-Kantarci,~\IEEEmembership{Senior Member,~IEEE}, Lajos Hanzo,~\IEEEmembership{Fellow,~IEEE}

\thanks{Roghayeh Joda and Medhat Elsayed were with the School of Electrical Engineering and Computer Science, University of Ottawa, Ottawa, ON, K1N 6N5, Canada when this research was conducted and they are currently with Ericsson Canada, Ottawa, ON, K2K 2V6, Canada  (e-mail:  roghayeh.joda@ericsson.com, medhat.elsayed@ericsson.com).}

\thanks{Melike Erol-Kantarci is with School of Electrical Engineering and Computer Science, University of Ottawa, Ottawa K1N 6N5, Canada (e-mail: melike.erolkantarci@uottawa.ca).}
\thanks{Akram~Bin Sediq and Gary~Boudreau are with Ericsson Canada, Ottawa, ON Canada (e-mail: akram.bin.sediq@ericsson.ca, gary.boudreau@ericsson.com). Hatem~Abou-Zeid and Ramy~Atawia were with Ericsson Canada, Ottawa, ON, K2K 2V6, Canada when this research was conducted. Ramy~Atawia is now with Dell Technologies, Ottawa, ON, K2B 8J9, Canada. Hatem~Abou-Zeid is now with the Department of Electrical and Software Engineering, University of Calgary, Calgary, AB, T2N 1N4, Canada (e-mail: ramy.atawia@queensu.ca,  hatem.abouzeid@ucalgary.ca).}
\thanks{Lajos Hanzo is with the School of Electronics and Computer Science, University of Southampton, Southampton, SO17 1BJ, UK (email:
lh@ecs.soton.ac.uk).}
\thanks{This work is partially supported by 5G ENCQOR program of Ontario Centres of Innovation and Ericsson Inc., Canada.  Melike Erol-Kantarci would like to acknowledge the support from Canada Research Chairs program.  Lajos Hanzo would like to acknowledge the financial support of the Engineering and Physical Sciences Research Council projects EP/W016605/1 and EP/P003990/1 (COALESCE) as well as of the European Research Council's Advanced Fellow Grant QuantCom (Grant No. 789028).}
}


\maketitle

\begin{abstract}
The Internet of Senses (IoS) holds the promise of flawless telepresence-style communication for all human `receptors' and therefore blurs the difference of virtual and real environments. We commence by highlighting the compelling use cases empowered by the IoS and also the key network requirements. We then elaborate on how the emerging semantic communications and Artificial Intelligence (AI)/Machine Learning (ML) paradigms along with 6G technologies may satisfy the requirements of IoS use cases. On one hand, semantic communications can be applied for extracting meaningful and significant information and hence efficiently exploit the resources and for harnessing a \textit{priori} information at the receiver to satisfy IoS requirements.
On the other hand, AI/ML facilitates frugal network resource management by making use of the enormous amount of data generated in IoS edge nodes and devices, as well as by optimizing the IoS performance via intelligent agents. However, the intelligent agents deployed at the edge are not completely aware of each others' decisions and the environments of each other, hence they operate in a partially rather than fully observable environment. Therefore, we present a case study of Partially Observable Markov Decision Processes (POMDP) for improving the User Equipment (UE) throughput and energy consumption, as they are imperative for IoS use cases, using Reinforcement Learning for astutely activating and deactivating the component carriers in carrier aggregation. Finally, we outline the challenges and open issues of IoS implementations and employing semantic communications, edge intelligence as well as learning under partial observability in the IoS context.    

\end{abstract}

\begin{IEEEkeywords}
Internet of Senses, semantic communications, edge learning, partial observability, 6G.
\end{IEEEkeywords}

\section{Introduction}
The vision of the Internet of Senses (IoS) in the year 2030 anticipates that users may become capable of flawless telepresence-style audio, video, tactile, olfactory and gustatory communications. In this vision, the physical and virtual reality become intricately intertwined. Thus, the IoS enables users to digitally convey the smell or the taste of food, texture of fabric, temperature of a surface as well as ultra high-quality video and audio. The IoS is considered to be a stepping-stone toward communicating thoughts using the Brain-Computer Interfaces (BCI) of the future\footnote{https://www.ericsson.com/en/reports-and-papers/consumerlab/reports/10-hot-consumer-trends-2030}. The main drivers for IoS research are online shopping, immersive entertainment, education and remote surgery, along with the virtual workplace that became an evident need due to the COVID-19 pandemic.  

Augmented, Virtual, eXtended Reality, Holographic-Type
communication, the tactile internet and digital twins are the use cases enriched by the IoS, which require high-rate, low-latency and high-reliability links \cite{Dohler2021,FG-NET2030}. These requirements, in unison with each other, allow us to enhance the true intimacy of flawless lip-synchronized stereoscopic or holographic telepresence. Please bear in mind that for gogle-free holographic IoS, different colored views must be conveyed at low latency. These IoS video clips are correlated and hence may be compressed, but sophisticated/complex holographic  compression imposes high latency. Clearly, these are complex trade-offs to be considered. In addition,
high-power computation capabilities are necessary for IoS use cases.  
Naturally, the high data rate as well as high computational and compression requirements typically lead to high energy consumption, hence energy efficiency becomes an issue for these emerging technologies. Moreover, the signals
conveying the representations of the different human senses and different streams have to be synchronized to prevent cybersickness \cite{Dohler2021}. Apart from the above-mentioned tight specifications, the human perceptual quality requirements also have to be considered in the IoS, which are related to human senses and cognition; and require innovation in the IoS devices. 
To achieve these ambitious goals, hyper-fast
and hyper-reliable connectivity, mobile edge computing
capability, context-awareness, Artificial Intelligence (AI), as
well as lightweight sense-actuating ‘gadgets’ are necessary, doing with human perceptual range and device level innovation.

Semantic communications is an emerging topic, especially in terms of context-awareness, where 
the meaning and the perceptual utility of  the messages conveyed are important \cite{ShiComMag2021}. Semantic communications has the potential of circumventing the classic Shannon capacity limits either by exploiting a \textit{priori} information at the receiver or by harnessing contextual information \cite{Nokia2021}.
It must be noted that this philosophy goes way beyond that of the joint source and channel coding of correlated sources.
In our vision, semantic communications, together with AI, may lead to significant advances in IoS. To achieve this superior performance, the receiver and the transmitter have to share a \textit{priori} knowledge (or a specific language and rules), which is a popular concept in ML. Hence, ML supports  semantic communications in extracting beneficial information. Additionally, semantic communications can be utilized for conveying messages between distributed ML agents relying on limited communication. 

Semantic communications and powerful 6G enabling technologies will support the IoS through enhancing both the data rate and reliability, while simultaneously reducing the latency \cite{SaadBannis2020}. Note that for satisfying these conflicting requirements, more bandwidth is necessary so as to provide the diversity gain or for accommodating  the increased redundancy of channel coding. Semantic-aware joint sensing, communication, reconstruction and control will be a challenge in IoS use cases. Additionally, to reduce the latency in the IoS, the computing resources must be placed closer to the edge and then harness ML algorithms for promptly providing services for the users. 
Furthermore,
ML identifies the pertinent knowledge in a deluge of data,
especially at the network-edge and maximizes the long-term
reward \cite{Mel2019}. Finally, ML is capable of solving complex Pareto optimization problems \cite{Hanzo2020}\footnote{The optimal Pareto front of a multi-component optimization problem is the collection of all optimal solutions, where for example neither the throughput, nor the Bit Error Rate (BER), nor the latency can be improved without degrading at least one of these metrics.}. Nevertheless, in the complex and especially distributed edge environment, the state of a specific intelligent agent might not be transparent to the others.
These partially observable states may be modeled by Partially Observable Markov Decision Processes (POMDP). However,
POMDP becomes excessively complex in multi-agent systems. Deep Reinforcement Learning algorithms play a fundamental
role in POMDP, but naturally, they also encounter challenges \cite{Nguyen2020}. Therefore, the success of IoS critically lingers on exploiting 6G  physical layer enabling technologies, semantic communications and edge intelligence. How intelligent agents at
the edge handle partial observability for overcoming these challenges also requires further research.

In \cite{ShiComMag2021} and \cite{Lan2021WhatIS}, the authors propose new solutions for semantic communications in the area of machine intelligence, while the latter also provides a holistic survey. The study in \cite{Nokia2021}  harnesses ML techniques for tuning the air-interface parameters. The authors in \cite{IDN2020} provide a comprehensive survey of intent-driven networks, where an application’s intent is translated to  a network-specific expression,  and then closed-loop control directly links the application’s intent to the network's configuration. To the best of our knowledge, our paper is the first one addressing the IoS and linking semantic communications, carrier aggregation and edge intelligence together as enablers of IoS, while also discussing the challenges of partial observability. Additionally, we unveil the intricate connection between semantic communication and ML algorithms, highlighting their beneficial collaboration. 

In the rest of the article, we first reveal some beneficial IoS use cases and their requirements in Section \ref{UseCases}. Then, the key 6G enablers of IoS are highlighted in Section \ref{KEYE}. We then critically appraise a case study that uses ML in the partially observable carrier aggregation scenario of Section \ref{CaseStudy}, as an example of enhancing the IoS data rate. Finally, Section \ref{OppChall} lists some of the critical challenges and open issues of developing IoS use cases in 6G networks.

\section{IoS Definition, Use Cases and Requirements}\label{UseCases}

\subsection{IoS Definition}
The IoS is defined as the notion of digitally streaming the signals conveying information for all senses and therefore, blurring the difference of physical and virtual reality as well as enabling telepresence style communications. Thus, as inspired by the Tactile Internet, the IoS architecture is presented as a 1) Primary domain (head-mounted VR glasses, haptic devices, holograms and BCI), 2) Network domain including edge computing, 3) Replica domain (audio, visual, olfactory, gustatory and tactile sensors and streams, haptic feedback, robots)  For instance, a consumer with his head-mounted VR glasses and haptic devices (primary domain and send the request) can sense the stuff with all senses in the grocery store (replica domain) which is equipped with sensors and cameras through the wireless network (network domain).


\subsection{Multi-sensory Extended Reality}
EXtended Reality (XR) includes AR, VR and merged  reality, all of which aim for creating realistic, interactive digital environments using sound and vision \cite{Samsung2020}. XR is extensively used in entertainment, education, medical applications and in the manufacturing industry. The next step of creating realistic environments involves using haptics, taste and smell to provide a truly immersive experience. Hence the retail, entertainment, education, health and food industries would be the first adopters of this alluring technology.  

There are ongoing efforts to replicate scents and tastes digitally. Digital scent technology, also known as olfactory technology, aims to sense, transmit and receive scent-enabled digital media. The sensing part of this technology \myuline{relies on} using olfactometers and electronic noses. After sensing, the digital flavor interface digitally stimulates taste (using electrical and thermal stimulation methodologies \myuline{applied to} the human tongue) and smell senses (using a controlled scent emitting mechanism) simultaneously \cite{SPENCE2017}. The transmission of end-to-end  olfactory and gustatory information using a wireless medium \myuline{imposes} \muline{tight throughput and latency specifications}. Additionally, the high-fidelity audio and video signals replicating the ambiance and augmenting human senses through XR interfaces further tightens the throughput and latency specifications. 


\subsection{Tactile Internet}
The tactile internet supports real-time access, monitoring and control of remote objects or processes using haptic feedback\cite{Dohler2021}. A typical example of the tactile internet use case is  remote surgery, where both the haptic sensory signals and audio-visual signals are transmitted from a remote site (or remote patient) to an operator (or a surgeon). Depending on whether the operator is directly interacting with the holograms or wearing \myuline{head-mounted} VR glasses, this audio-visual monitoring can be provided by immersive 3D HTC or XR video streaming techniques \cite{FG-NET2030}. Based on the information obtained from different inputs (haptic and audio-visual), the operator then transmits the corresponding control signal to the remote site. The tactile internet has been a hot research topic for years and it is a compelling use case of IoS.     

\subsection{Holographic-Type Communication}
Holographic-Type Communication (HTC) is based on digitally delivering 3D images between remote locations \cite{FG-NET2030}. It relies on a multitude of cameras recording the same scene from equi-angle positions spread across 360°, where remote users appear with their holographic presence. Holographic displays have to satisfy all  aspects of human observation such as viewpoint tilts, angles and positions relative to the hologram in addition to the color, depth, resolution and frame rate of standard video specifications \cite{Dohler2021}. 
HTC will play an important role in IoS, as it creates realistic telepresence environments. Naturally, the multiplicity of angular views are highly correlated, hence they lend themselves to high compression holographic encoding, which is in its infancy and requiring further research.

HTC is naturally related to human vision, but when further augmented with data from other senses, it can facilitate the creation of a digital twin of an environment as highlighted in the following section.

\subsection{Digital Twin}
A digital twin is the virtual representation of physical (real) entities, e.g., persons, objects, \myuline{processes} or systems 
\cite{FG-NET2030}. By relying on a digital twin, users may become capable of real-time  monitoring and  controlling the physical reality in a virtual environment. 
In general, XR technologies or holograms are utilized to observe and explore the virtual world with the aid of 3D imaging.
DTs are already employed in manufacturing and industrial applications. However, in the future IoS, they can
be augmented with the aid of multi-sensory inputs for facilitating many
compelling applications \cite{khan2021digitaltwinenabled} such as online learning and education.  

\subsection{Key Network Requirements}
The communication requirements of 6G use cases have been identified in  \cite{Dohler2021,FG-NET2030,Samsung2020}, some of which are applicable to the IoS.
According to these reports, immersive 16k VR requires 0.9 Gbps throughput assuming  a compression ratio of 400. 
For the tactile internet, the required throughput varies, depending on the motion-activity of the specific application. True holograms and computer-generated holograms require throughputs up to Terabits/sec (Tbps). For example, a hologram of 19.1 Gigapixel resolution requires 1 Tbps. Thus, at least a rate of  0.58 Tbps is required for conveying a hologram of 11.1 Gigapixel resolution over a mobile device. Thus, HTC needs ultra-high rates and bandwidth. However, holographic video-compression coupled with activity-dependent non-uniform sampling has the potential of mitigating these challenges. Finally, virtualized objects have to be represented by a huge amount of data. For instance, a digital twin of a  $1 \times 1$ $m^2$ area requires 0.8 Tbps throughput at a compression ratio of  300 for the first few frames, although subsequent updates of the digital twin may require lower data rates.


Latency is also a key performance indicator, and the critical applications of the IoS in remote surgery require ultra-low latency at a sub-ms level and ultra-high reliability \cite{FG-NET2030}. Moreover, low latency is required for preventing cyber-sickness. For example, 360° VR systems have to react to head movements with a latency below 10 ms.
The latency, reliability and security  requirements are very stringent in critical real-time communications. Moreover,  reliability is a salient factor in certain applications that demand deterministic performance. Synchronization among different sensory inputs or multiple transmission paths or data streams is also necessary in IoS use cases. Furthermore, power, storage and computation resources are required at the edge for HTC use cases and for frame synchronization, and for  synthesizing, rendering and reconstructing 3D images.

\begin{figure} 
    \centering		
    \includegraphics[width=3. in]{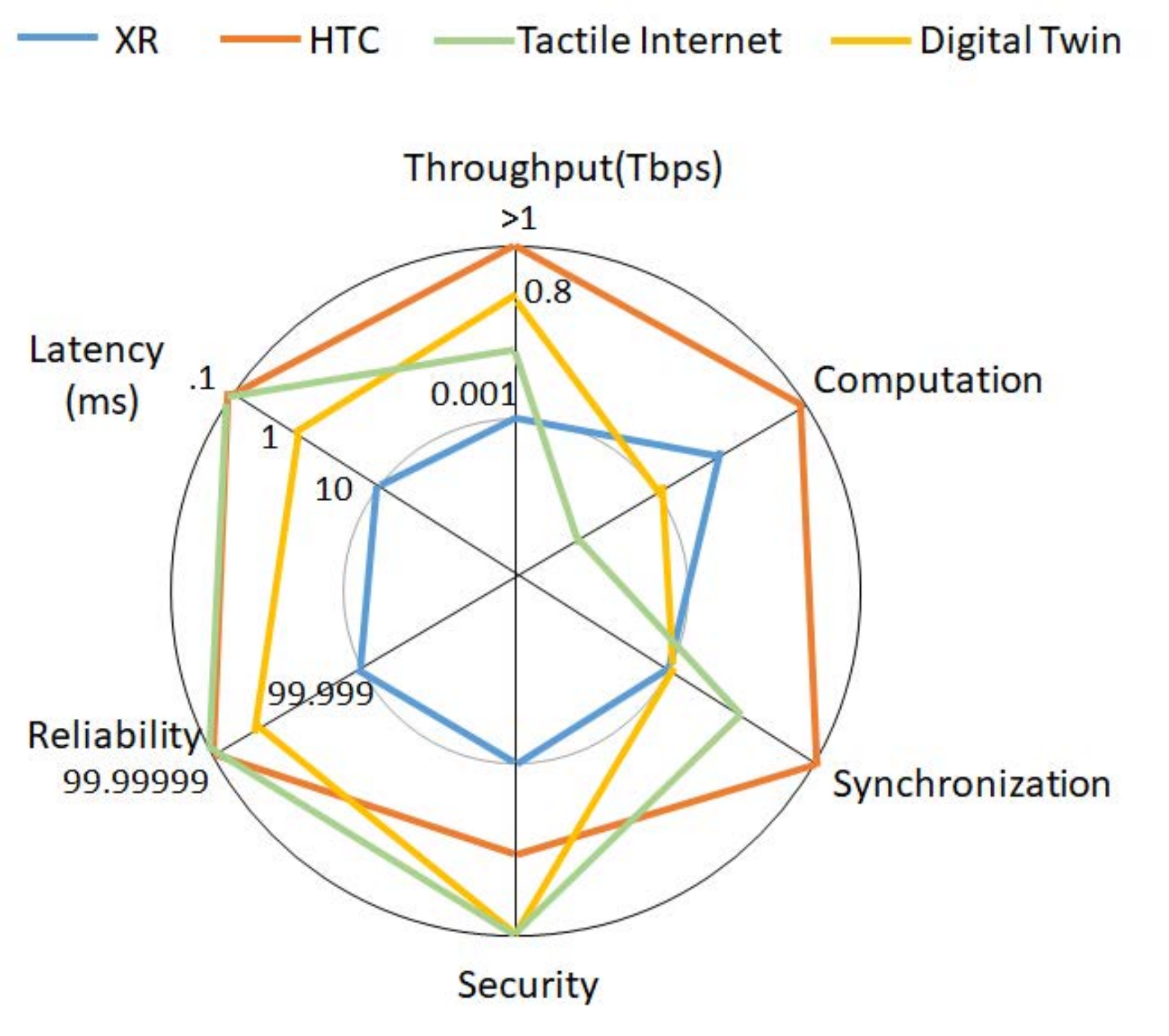}
		\caption{Network requirements for IoS use cases. If HTC, digital twin or XR are used for critical applications, they should satisfy the high-reliability and low-latency requirements. Additionally, if the tactile internet exploits other use cases, it has to meet their corresponding requirements. Note that the IoS use cases constitute a subset of 6G use cases; and according to the applications and services relying on the IoS, one of these use cases or a combination of them is employed.}
		\label{ReqT}
\end{figure}

\begin{figure*} 
    \centering		
    \includegraphics[width=7 in]{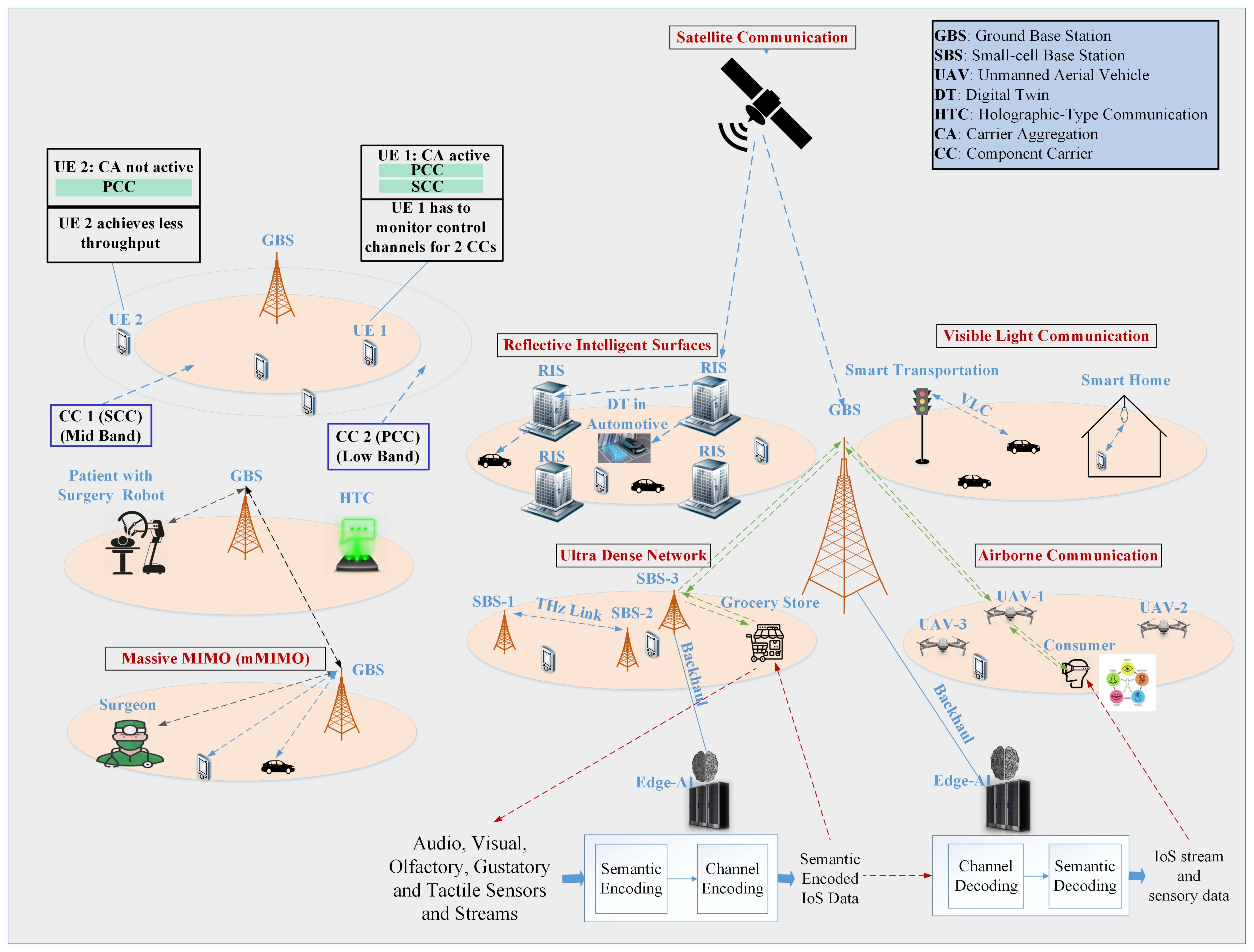}
		\caption{Vision of a 6G network supporting IoS use cases including a grocery store connected to an intelligent edge server equipped with semantic encoding and decoding, remote surgery robots and patients connected to a surgeon, a HTC scenario and the digital twin of a car. For instance, grocery store sends its sensory and stream information to the edge server and output semantic encoded data is conveyed to the edge server supporting the consumer.  Semantic encoding includes semantic sampling, filtering, stream combining and feature extraction. Note that red dash lines are virtual links. 
		In carrier aggregation shown, one extra Component Carrier (CC)  as the secondary CC (SCC) can be activated in addition to the primary CC (PCC).}	
		\label{IoSModel}
\end{figure*}

\myuuline{The above-mentioned requirements of IoS use cases are shown in Fig. \ref{ReqT},} while fledgling 6G technologies and IoS use cases are illustrated in Fig. \ref{IoSModel}. IoS use cases are empowered by exploiting a variety of 6G technologies to fulfill their requirements. The figure features powerful 6G technologies, such as
RIS, massive MIMO, SAGIN \cite{SaadBannis2020}, carrier aggregation and visible light communication \cite{Dohler2021}. 
The figure also shows a grocery store example that is connected to an intelligent edge server equipped with a semantic encoder and decoder, remote surgery robots, an HTC scenario and the digital twin of a car. 

\section{Key Enablers of IoS in 6G}\label{KEYE}

\subsection{Semantic Communications}\label{SC}
Technological advances in communication over the past decades have predominantly aimed for achieving reliable high-rate transmission. However, semantic communication additionally considers the meaning, significance and utility of the  information conveyed \cite{Lan2021WhatIS}. Furthermore, by considering a priori information at the receiver,  higher throughput can be achieved for IoS use cases. \myuuline{Specifically, in human-computer interaction applications, such as the digital twin of mobile robots and surveillance  of an environment, the sampling, transmission and control of information may be triggered, whenever a change occurs at the source}. Then, the receiver may apply semantic reception for decoding the information, again, according to its usefulness. Additionally, data analytics and ML may be utilised for making decisions concerning the usefulness of information based on past experiences.      

\myuuline{Semantic information can be extracted based on different features such as the value of information, age of information, timeliness and relevance} \cite{Lan2021WhatIS}. \myuuline{Semantically-aware filtering can be applied at the source to acquire the relevant samples of signals for transmission based on their freshness and significance}. This may reduce the sampling rate below the Nyquist-rate, which reduces the channel occupancy. Furthermore, semantically-aware preprocessing could be applied to extract important features for transmission. For example, \myuuline{in the virtual workplace, some features of the environment, such as scent, can be inspected first and only the relevant information would be transmitted} instead of sending all the acquired information. Then, the matched receiver applies semantic reception to reconstruct the samples in real-time based on its target quality. In summary, semantic communications motivate the joint design of sampling, transmission, reconstruction and \myuuline{the control} of information, while considering the associated source variations, reconstruction efficiency, channel parameters, and QoS requirements, such as delay, throughput and reliability.

Furthermore, the interplay between semantic communications and ML \muline{may be exploited} for improving both \myuline{the communications} and learning metrics \cite{ShiComMag2021, Lan2021WhatIS}. For instance, \myuuline{in IoS, imagine} two distributed \muline{cell-edge} nodes, \myuline{where one of them} senses the environment and \muline{the other}  selects \myuline{a bespoke} action according to this sensed information. However, the  information has to be  transmitted over \myuline{imperfect channels to} the second node. Here the objective of the communication protocol (e.g., modulation and coding parameters) is not only \muline{that of} increasing the reliability, but also achieving the learning goal of the second node. This is a typical example of semantic communication. Thus,
semantic communication is an emerging field of study and has \myuline{the potential of supporting} IoS applications in 6G.

\subsection{Emerging Physical Layer Techniques}\label{KEYE-PHY}
Naturally, increasing both the spectral efficiency and the available bandwidth improves the network capacity\myuuline{, which is a key requirement of IoS use cases.} A popular technology is constituted by RISs relying on passive reflectarrays and control elements for minimizing the \muline{BS's}  power  consumption \cite{SaadBannis2020}.
Moreover, exploiting \muline{the large bandwidth of the mmWave and THz bands allows} augmenting the throughput\myuline{, but at the cost of high pathloss, which must be compensated by high-gain beamforming. They \muline{also} necessitate sophisticated} mobility management techniques. Nevertheless, \myuline{the sub-6GHz frequency bands are still essential for improving} the coverage.  
\myuline{Additionally, the SAGIN concept is \muline{pivotal} for providing ubiquitous coverage, since it is capable of filling the coverage holes}. In addition to the above-mentioned advances, multi-connectivity and carrier aggregation continue to be important capacity-enhancing techniques. Although carrier aggregation has been used since LTE, there are still challenges that have to be addressed. For instance, the UE's power consumption increases with carrier aggregation, because all active component carriers are monitored even though they are not used for data transmission. Furthermore, there are numerous challenges in carrier aggregation in dual connectivity scenarios and in the allocation of bandwidth to control channels. Carrier aggregation will also be an important technology for IoS, but erratic fluctuations in traffic, strict QoS requirements and several other factors will make it difficult to reach optimal decisions under uncertainty. Therefore, there is a need for learning carrier aggregation decisions in partially observable environments.

\subsection{Edge Intelligence and Learning in Partially Observable Environments}\label{EdgeI}
The growing number of edge servers provides abundant computing resources distributed at the network edge. 
Naturally, edge intelligence continues to be a key-enabler for the IoS, as it can be harnessed for the optimal and proactive exploitation of communication and computing resources that leads to improved throughput, reliability and latency which are beneficial for IoS use cases. In fact,  AI will play a prominent role in the design, deployment and  operational phases of 6G networks \myuuline{and IoS}. 
Since the network will be flexible and programmable, AI facilitates automated network optimization and management, and as an additional benefit, a variety of tasks including radio resource management, \myuuline{carrier aggregation}, mobility management, virtualized network function placement and slicing can be improved by exploiting AI. Additionally, designing the air interfaces by ML algorithms under AI-native air interfaces may enhance the wireless transmission \cite{Nokia2021}. 
Moreover, UE can be trained to learn both protocol parameters as well as control messages to optimize the network performance.

In ML-aided wireless systems, it is routinely assumed that intelligent agents are capable of fully observing the environment. However, in practice, there is substantial uncertainty in the state of the environment encountered, hence the agents may have to make decisions in the face of partial observability. The popular POMDP technique is eminently suitable for handling such \myuuline{circumstances}. \myuuline{In particular,} distributed learning in edge-intelligence typically relies on access to partial information at each edge node. Communicating in the face of limited information about the environment becomes even more challenging, when there is limited communications among the nodes. Therefore, distributed learning at the edge has to be modeled by a Multi-agent Markov Decision Process (MMDP) relying on partially observable state information, which is also known as Decentralised POMDP (DEC-POMDP) or Multi-Agent POMDP.

\begin{figure} 
    \centering		
    \includegraphics[width=3.in]{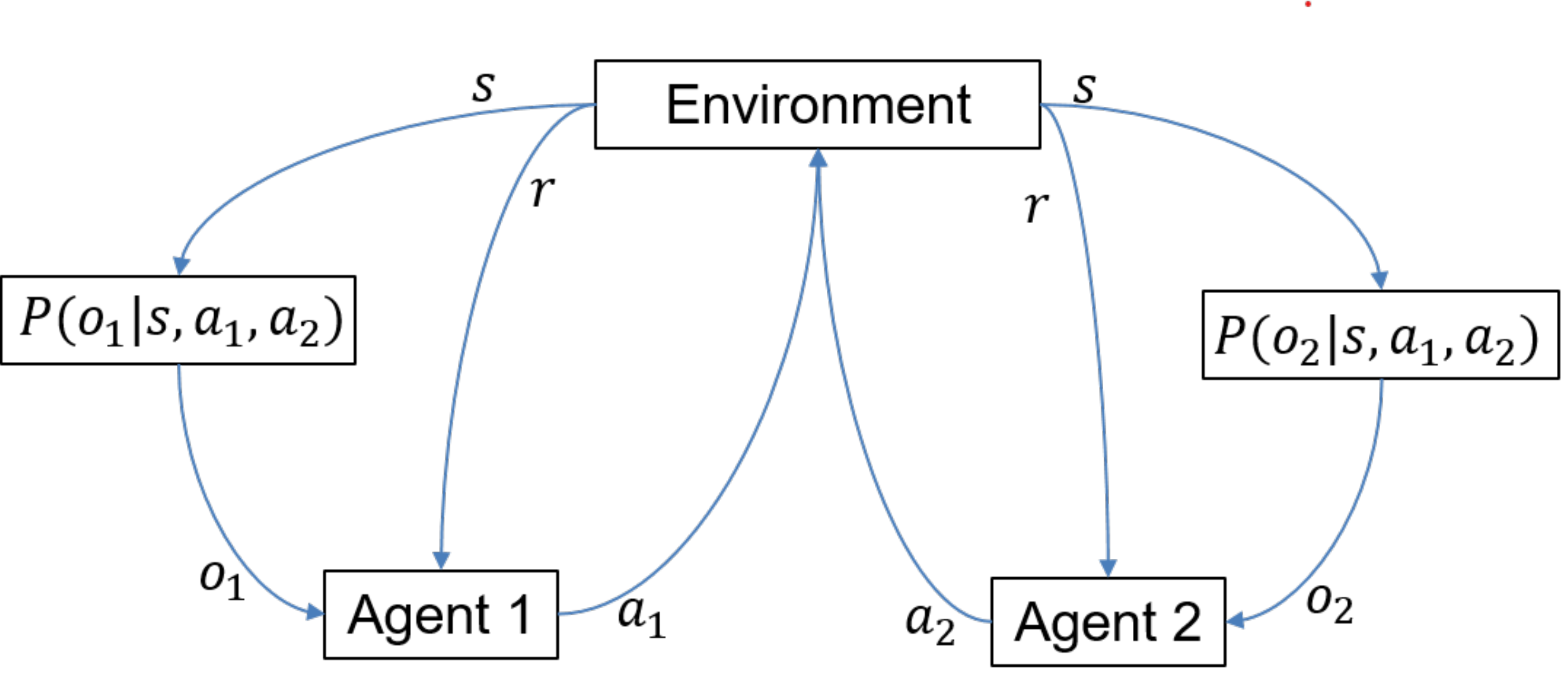}
	\caption{POMDP model with two agents, where each agent accesses its own  observation. The observation for each agent depends on the sate of the environment and the action selected by both agents.}	
	\label{DEC-POMDP-Model}
\end{figure}

A distributed DEC-POMDP model relying on a pair of agents is portrayed in Fig. \ref{DEC-POMDP-Model}, where the environmental state $s$ is observed as $o_1$ and $o_2$ by agents 1 and 2, respectively. These observations depend both on the previous actions carried out and on the resultant state, as well. The pair of agents select the appropriate actions $a_1$ and $a_2$ according to the POMDP policy considered. The actions chosen will then alter the current state to a new state of the environment. Correspondingly, this updated state creates new observations for the agents. An intelligent POMDP policy acts as a function of the previous observations and actions, but additionally it also depends on the instantaneous reward achieved. 
A pivotal example of partial observability in the wireless domain emerges in the context of carrier aggregation, \myuuline{given the uncertainty in terms of the traffic arrival statistics}. In the next section, we provide deeper insights into this problem, since efficient carrier aggregation is crucial for enhancing the throughput \myuuline{or equivalently transmission delay} in IoS use cases.
 

\section{Case Study on Carrier Aggregation under Partial Observability for IoS Applications }\label{CaseStudy}
\myuuline{Again, carrier aggregation} is one of the key enabling techniques for IoS, where the users can aggregate multiple CCs by activating Secondary CCs (SCCs) in addition to the Primary CC (PCC). Although persistent activation of all SCCs improves the throughput, it increases the energy consumption, owing to the continuous monitoring of all CCs by the UEs. Therefore, a dynamic carrier aggregation algorithm is required for promptly activating and deactivating the SCCs at traffic arrivals and departures, respectively. While this reduces the average number of active CCs, it should maintain a high throughput similar to that keeping all CCs active simultaneously.  

\myuuline{A wireless network of one gNB with a cell radius of 250m is considered, where the  users are uniformly distributed. The gNB has three CCs and each CC has a 5 MHz bandwidth. The traffic is categorized into two classes: File Transfer Protocol (FTP) and Constant Bit Rate (CBR) traffic.
However, the traffic pattern of users is unknown to the carrier aggregation manager. Thus, the carrier aggregation manager aims for optimizing CC allocations under partial observability}. In our case study, we use model-free Q-learning for SCC activations and deactivations according to the traffic estimation process. In particular, the Q-learning state is determined by three parameters: the estimated inter-arrival time, the estimated data size and the average throughput.  
Furthermore, the actions are the activation and deactivation decisions of the available SCCs. Finally, the reward is formulated by assigning credits to those actions that activate SCCs which match data arrivals and additionally satisfy data size requirements. On the other hand, the reward penalizes activation decisions that result in wasted CC allocations. 

\begin{figure*}
\centering
 \hspace*{-.5cm}
     \includegraphics[width=6.in]{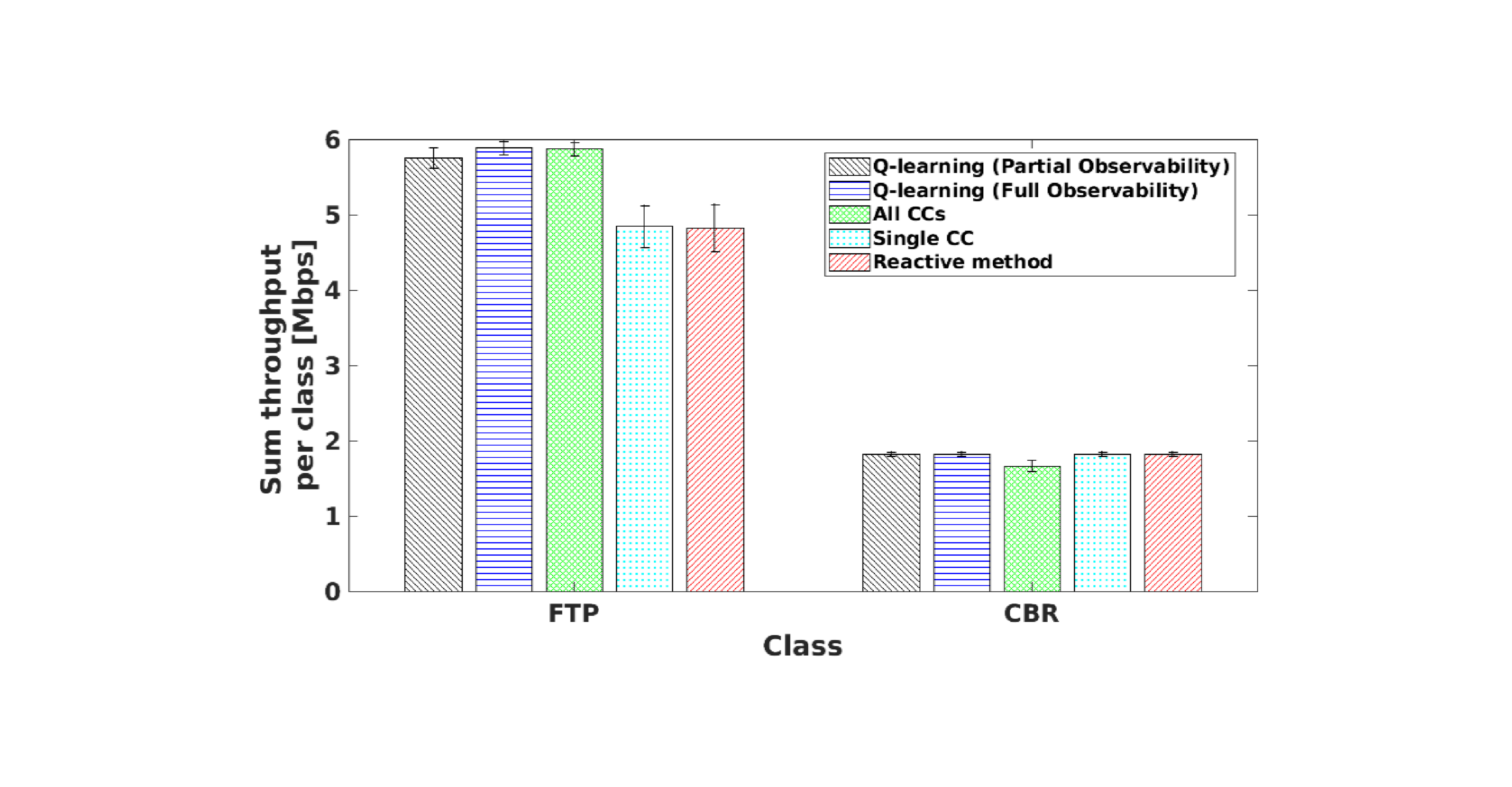}
     \vspace*{-1.cm}
    \caption{Sum Throughput Per Class of Users.}
    \label{fig:sumTpt}
\end{figure*}

\begin{figure*}
 \centering
 \hspace*{-1.cm}
    \includegraphics[width=6.in]{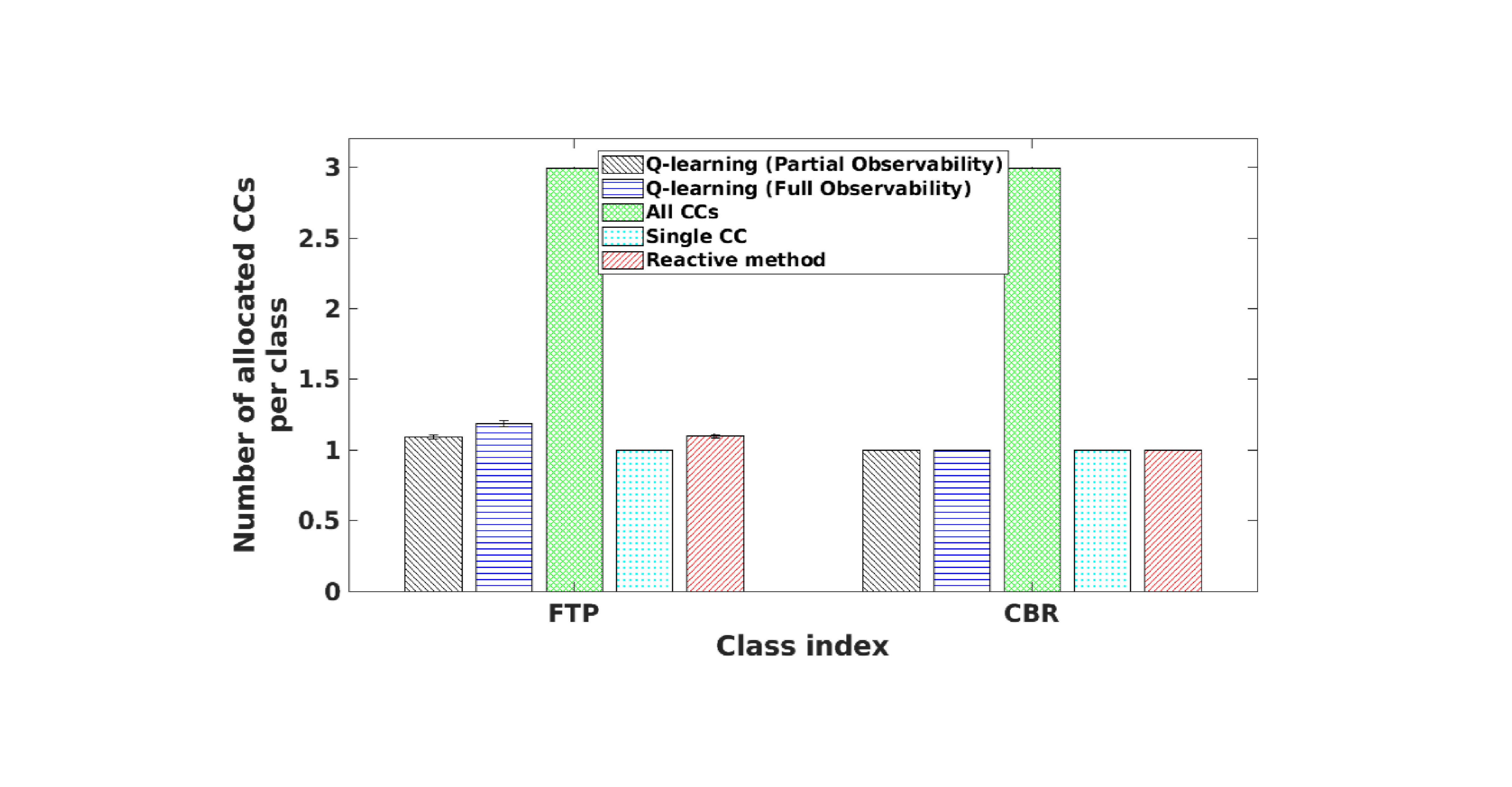}
    \vspace*{-1.6cm}
    \caption{Average Number of Allocated CCs Per Class of Users.}
    \label{fig:numCcs}
\end{figure*}

Figs. \ref{fig:sumTpt} and \ref{fig:numCcs} present the sum throughput and the average number of active CCs for five algorithms. The methods compared are as follows. `All CCs' method always activates all SCCs, whereas `Single CC' uses only \muline{one} PCC per user. The `Reactive method' only activates/deactivates SCCs after confirming data arrival/departure respectively which leads to reduced throughput, as shown in Fig. \ref{fig:sumTpt}. Finally, `Q-learning' \myuline{associated} with full observability assumes the \myuline{perfect} knowledge of the \myuline{`Q-learning'} state (i.e., arrival time and data size)\myuline{, while `Q-learning' with a partially observable state,} takes uncertainty into consideration. 

The figures demonstrate the ability of `Q-learning' to dynamically activate and deactivate SCCs \myuline{upon} data arrival, hence achieving \muline{a similar} throughput to \myuline{`All CCs'} without incurring a large number of active CCs. \myuuline{It is noteworthy that the UE energy consumption increases as the number of activated CCs is augmented} \cite{Jodaletter}. On the other hand, the `Reactive method' is able to reduce the number of allocated CCs, but, it fails to achieve high throughput due to the delay of waiting for the next activation-deactivation decision. Although PCell is active by default, a decision to activate and deactivate SCCs is performed periodically and according to data arrival/departure. Finally, `Q-learning' with partially observable states is capable of performing similarly to that in a fully observable environment.  {However, `Q-learning' with partial observabililty convergences after about 600 time slots, which is nearly $80\%$ slower than `Q-learning' with full state knowledge, due to requiring more exploration time. Hence, we can train the agents of the proposed algorithm in an offline mode and then use the trained model during normal operation and operating mode either during execution or when the inference is made.}


\begin{table*}
\caption{\label{TChallenges}IoS Challenges and Open Issues}
\begin{tabular}{ |p{5cm}|p{1.7cm}|p{10.cm}| }
\hline
 \textbf{Open Issues} &\textbf{Categories}& \textbf{Challenges} \\ &&\\
\hline
   Human  Perception Range and  Device-Level  Innovations&Device-layer challenge& Considering human perception range in IoS device design  and data transmission techniques\\
  \hline
 Multi-Sensory  Input  Synchronization  and  Coordination &Device-layer challenge& Synchronization and coordination among different data sources   and feedback data to reduce cyber-sickness\\
 \hline
 Compression Schemes &Device-layer challenge&Reducing the data size and designing \myuline{compression techniques suitable} for IoS use cases  \\
  \hline
  Edge Learning under Partial Observability&Cross-layer challenge&Design the objective-oriented and semantically-aware learning algorithms at the edge, while the environment is not completely known for the edge nodes\\
\hline
 Joint Sensing, Communication, Reconstruction and Control&Cross-layer challenge& Jointly designing sensing, communication schemes along with reconstruction and  control at the receiver exploiting ML algorithms to increase the efficiency under state uncertainty\\
 \hline
  Security  and  Privacy&Cross-layer challenge& Design  secure communication protocols adaptable to a variety of IoS use cases and  \myuline{environmental status. Additionally}, preserve the   data privacy for IoS use cases\\ 
  \hline
 \end{tabular}
\end{table*}

\section{Challenges and Open issues}\label{OppChall}
In this section, we highlight the challenges and open problems related to the IoS devices and to the enabling technologies.  The challenges are divided into two categories:  a)
      IoS device-layer challenges and
   b) IoS cross-layer challenges. Note that IoS devices tend to suffer from battery charge limitations and many of the above-mentioned enablers, including carrier aggregation, typically increase the energy consumption of the device  \cite{Jodaletter}. Also, for rendering and reconstructing 3D images in HTC, high computation power is necessary and therefore, it is inevitable to offload parts of the demanding computing tasks to the edge servers. For these challenges, we refer the reader to \cite{Jodaletter} and \cite{FG-NET2030}.
\subsection{IoS Device-Layer Challenges}
\subsubsection{Human Perception, Range and Device-Level  Innovations for  IoS}
It is \myuline{exploited} that in 360° video, it is sufficient to transmit only \myuline{a limited} field of view, since humans can only focus \muline{on} a limited segment of a 360° surround video, while being oblivious \muline{of} details in the background. 
\myuline{Limited ranges in human perception may also} exist for taste, smell and haptics. This \myuline{has} to be investigated, when \myuline{new} codecs are developed and intelligent algorithms may also be conceived for perceptual \muline{weighting} of a scent differently from others. For instance, a light odor that is masked \myuline{by strong ones} may not be transmitted. Moreover, it is \myuline{routinely exploited} that human sight detects a small range of wavelengths, and that people do not hear any sound above 15-20 kHz. 
On the other hand, the IoS provides enhanced levels of human perception and sensing, say in treating illnesses or controlling a vehicle. 
Thus, \myuuline{the user is at the center of IoS transceiver design}.

\subsubsection{Multi-sensory Input Synchronization and Coordination}
The streams of video, audio, haptic, olfactory,  gustatory data and 3D images taken from \myuline{multiple} viewpoints have to be synchronized \myuline{for avoiding} cyber-sickness. \myuline{Additionally}, the streams \myuline{arriving} from different positions and disparate paths \myuline{have to be tightly} coordinated for providing an immersive IoS experience in 6G.           

\subsubsection{Compression Schemes}
\myuline{Uncompressed} Holograms in HTC require \myuline{excessive} bandwidth. \myuline{Therefore}, efficient data  compression techniques are \muline{required} for reducing the data rate. \myuline{Additionally}, the kinesthetic data, tactile data in tactile internet along with \myuline{the} audio-video streams \myuline{of} XR  or 3D images \muline{have} to be frequently \myuline{refreshed for maintaining high} performance. To minimize the feedback rate, \myuline{the feedback} has to be compressed at the remote site and then transmitted \muline{back} to the operator.

Data \myuline{compression} techniques \myuline{typically impose variable delay on the feedback. The feedback \muline{channel} imperfections impose future} challenges on the management and synchronization of data streams  \cite{Dohler2021}. Thus, \myuuline{developing} appropriate compression techniques \muline{for} multi-sensory IoS use cases are among the \myuline{critical} open research areas.

\subsection{IoS Cross-Layer Challenges}
\subsubsection{Edge Learning under Partial Observability}
\myuuline{Distributed nodes at the edge should learn to efficiently allocate the resources for IoS use cases according to the specific application requirements, optimization objectives, the environment and hardware limitations. Obviously, edge learning is under partially observable conditions, which encounters numerous challenges. One approach to reduce uncertainty is to collaborate and share information among nodes. However, the communications between nodes are imperfect. Thus, 
semantically-aware and goal-oriented communications for learning among collaborating edge nodes under partial observability has to be investigated} \cite{GunduzMag2020}.

\subsubsection {Joint Sensing, Communication, Reconstruction and Control Design in IoS}
Applying source sampling \myuline{based} on the semantic knowledge extracted from  intelligent devices (e.g. updated information in XR use cases) and then employing channel coding \myuline{independently} of the  knowledge \myuline{obtained} is a sub-optimal solution. If we utilize the best communication schemes \myuline{for transmitting obsolete} (or stale) information produced at the source, we in fact waste the communication resources \cite{Lan2021WhatIS}. \myuline{A more efficient approach is to reconstruct} the source samples at the receiver and \myuline{then opt for the most appropriate} control actions \myuline{for joint} source sensing and communication. \myuuline{For example, low-activity video-scenes require lower frame-rates than a sudden burst of high-dynamic activity}.
Explicitly, semantically-aware joint sensing, communication, reconstruction and control needs to be investigated in more details for improving the performance of IoS.

\subsubsection{Security and Privacy} 
Security and privacy play important roles in IoS applications. For instance, \muline{a cyber-attack} during a remote surgery may have \myuline{fatal} consequences. Likewise, the digital twin of a person would carry valuable personal information that \myuline{has} to be kept private. \myuline{Additionally}, since digitised data is distributed in edge-servers, \myuline{computations have} to be  confidential.
Thus, secure and privacy-aware techniques have to be developed for IoS applications. 

We summarize the \myuline{above} challenges and open issues \myuline{at a glance} in Table \ref{TChallenges}.

\section{Conclusion}
In this paper, we discussed the IoS, including its use cases and communication requirements. To fulfill IoS requirements, we \myuline{elaborated on} semantic communication, edge intelligence and \myuuline{the emerging 6G} technologies. \myuline{Given} these technologies, we discussed how uncertainty and partial observability became important paradigms for ML algorithms. To showcase the importance of partial observability, we focused on a carrier aggregation case study, where secondary carriers are dynamically activated and deactivated with the objective of improving both the throughput and latency, while simultaneously considering the energy consumption. Finally, we discussed the challenges and open problems of IoS applications.



{}

\end{document}